\documentclass{article}

\usepackage{booktabs} %
\usepackage{fancyvrb}
\usepackage{tikz-cd}
\usepackage{algorithm}
\usepackage{enumitem}
\usepackage{algpseudocode}
\usepackage{bm}
\usepackage{amsmath,amsthm, amssymb}
\usepackage{balance}
\newcommand{\eat}[1]{}

\newtheorem{theorem}{Theorem}
\newtheorem{corollary}[theorem]{Corollary}

\newtheorem{lemma}[theorem]{Lemma}

\theoremstyle{definition}

\newcommand{\defn}{:=}
\DeclareFontFamily{U}{mathx}{\hyphenchar\font45}
\DeclareFontShape{U}{mathx}{m}{n}{
      <5> <6> <7> <8> <9> <10>
      <10.95> <12> <14.4> <17.28> <20.74> <24.88>
      mathx10
      }{}
\def\cov{{\rm Cov}}
\def\E{{\ensuremath{\mathbb E}}}

\newcommand{\eqdist}{\ensuremath{\stackrel{d}{=}}}

\newcommand{\convp}{\ensuremath{\stackrel{p}{\to}}}
\def\convd{\rightsquigarrow} % convergence in distribution
\long\def\comment#1{}

\newcommand{\sign}{\ensuremath{{\rm sign}}}

\newcommand{\acal}{\ensuremath{\mathcal A}}

\newcommand{\ccal}{\ensuremath{\mathcal C}}
\newcommand{\dcal}{\ensuremath{\mathcal D}}

\newcommand{\fcal}{\ensuremath{\mathcal F}}
\newcommand{\gcal}{\ensuremath{\mathcal G}}

\newcommand{\ical}{\ensuremath{\mathcal I}}

\newcommand{\pcal}{\ensuremath{\mathcal P}}

\newcommand{\tcal}{\ensuremath{\mathcal T}}

\newcommand{\vcal}{\ensuremath{\mathcal V}}

\newcommand{\xcal}{\ensuremath{\mathcal X}}

\newcommand{\change}[1]{{#1}}

\graphicspath{ {./figs/} }
\begin{document}

\title{Adaptive threshold sampling}
\author{Daniel Ting \\ dting.tr@gmail.com
\thanks{This work was done while the author was employed at Tableau Software.}}

		\maketitle

	\begin{abstract}		
		Sampling is a fundamental problem in computer science and statistics. However, for a given task and stream, it is often not possible to choose good sampling probabilities in advance. 
        We derive a general framework for adaptively changing the sampling probabilities via a collection of thresholds.
        In general, adaptive sampling procedures introduce dependence amongst the sampled points, making it difficult to compute expectations and ensure estimators are unbiased or consistent. Our framework address this issue and further shows when adaptive thresholds
        can be treated as if they were fixed thresholds which samples items independently.
        This makes our adaptive sampling schemes simple to apply as there is no need to create custom estimators for the sampling method.

        Using our framework, we derive new samplers that can address a broad range of new and existing problems including sampling with memory rather than sample size budgets, stratified samples, multiple objectives, distinct counting, and sliding windows. In particular, we design a sampling procedure for the top-K problem where, unlike in the heavy-hitter problem, the sketch size and sampling probabilities are adaptively chosen.

	\end{abstract}

	\section{Introduction}
	Sampling is a fundamental problem in  computer science and statistics. 
	By reducing the amount of data processed, it can significantly improve performance and lower costs, or it can ensure that the data processed fits within a system's resource constraints. 
	Of particular interest are random sampling without replacement procedures as they do not sample redundant information. 

	Before observing the data, it is not possible to choose appropriate sampling probabilities. For example, given a weighted stream of unknown length, it is \change{impossible} to choose a sampling probability ahead of time \change{that ensures the sample satisfies a finite memory budget $B$}.
	To guarantee \change{a finite} budget is satisfied, each item's inclusion decreases the available budget and affects the inclusion probability of other items. 

	This dependence causes several difficulties when designing sampling procedures and deriving estimators. In particular, the dependence often makes it intractable to compute the inclusion probability for each item or sets of items. When these sampling probabilities cannot be estimated, the sample is almost useless in data analysis. Good estimates cannot be obtained since estimators must adjust the contribution of each item based on its inclusion probability. 

	These difficulties can extend to designing sampling procedures. When the sample size is not fixed, items can be drawn independently with weight 
	$w_i$ with probability $\pi_i \propto w_i$. For example, the Conditional Poisson Sampling scheme is one that draws a fixed size sample and has desirable properties; however, no known algorithm can efficiently draw Conditional Poisson samples.
	
	We propose a framework, \emph{adaptive threshold sampling}, that addresses all of these challenges. We use it to solve novel problems and improve existing solutions. In this framework, samples are easy to draw; sample sizes and probabilities can change on the fly; and good estimators can be derived even though samples are dependent. This framework mimics drawing independent (Poisson) samples.
	Each item $x$ is associated with an independent random value $R_x$ and a threshold $T_x$. The item is included in the sample if $\change{R_x} < T_x$. 
	\change{Rather than choosing a constant threshold $T_x$}, we adjust the threshold in a data and sample dependent way to obtain desirable properties.
	
	Our methodological contributions revolve around making it \change{simple} to build thresholds where the resulting sample is easy to analyze. 
	We establish conditions when the threshold $T_x$ can be treated \emph{as if} it was a fixed threshold that yields an independent sample. 
	This simplifies analysis of the sample since one can apply an existing unbiased estimator for independent samples.
	Deriving and analyzing a custom estimator based on the true sampling distribution becomes unnecessary.
	When the conditions do not hold, we introduce a more general notion of \emph{threshold recalibration} that makes it easy to compute expectations and derive new unbiased estimators for a broader class of thresholding rules.
	We also provide methods for building and composing thresholds and for merging samples.
	We also prove an empirical process convergence result that further simplifies the development of good estimators and sampling designs.
	It extends our theory for unbiased estimators and shows when consistent estimators for independent samples remain consistent when applied to adaptive thresholding samples. It also provides justification for heuristic thresholding schemes 
	that do not satisfy our conditions for unbiased estimation but satisfy an asymptotic convergence condition which may be easier to verify.

    Our contributions to applications exploits our
    methodological contributions to develop new or improved sketches and estimators on a range of new and existing problems.
    For example, we double the effectiveness of the state-of-the-art in sampling from sliding windows \cite{gemulla2008sampling} even though we use exactly the same sketch to construct the sample. We improve merge procedures for distinct counting sketches, handle budget constraints for samples given variable length items, 
    and provide a solution to a novel top-k problem, a more challenging variation of the well-studied heavy hitter problem, where the top-k items by frequency must be returned regardless of how small their frequency may be.

	\vspace{-0.1cm}
	\subsection{Related work}
	    A long line of work has studied the bottom-k sample
	    \cite{rosen1997asymptotic, rosen1997sampling, duffield2007priority, cohen2007bottomk}, showing how it can be used to draw weighted samples \change{while} deriving good estimators for sums \change{over subpopulations} as well as estimates of their variance. 
	    These bottom-k sampling methods can be considered an adaptive threshold sampling method where the threshold ensures the sample size is exactly $k$.
	    Extensions \cite{cohen2009leveraging, cohen2015multi} study combining bottom-k samples. \change{Another techniques for efficiently drawing fixed size $k$ samples is VarOpt sampling \cite{cohen2009stream}.} 
	    
	    Little work has \change{provided generalizations of} the bottom-k thresholding rule
	    while providing a unbiased estimators. 
	    \change{One notable exception is} in distinct counting applications using $\mathit{Uniform}(0,1)$ priorities. 
		\change{The} Theta sketch \cite{dasgupta2016thetasketch} provides a general 1-goodness condition. Thresholding rules that satisfy this ensure cardinality estimates are unbiased. For uniform sampling, a generalized thresholding rule has been used for distributed sampling \cite{tirthapura2011distributedreservoir}.

\change{In addition to}
 providing an adaptive sampling framework, \change{our contributions also use} it to address novel problems and improve existing methods.
    Novel problems include sampling with fixed memory budgets, multi-stratified sampling, and top-k queries. These problems are more difficult
    variations of existing problems. 
    Examples of existing sketches and sampling methods that are improved include
	    sampling in sliding windows  \cite{gemulla2008sampling} and distinct counting \cite{giroire2009order, ting2016towards, dasgupta2016thetasketch}. 
	    Our applications are provided as examples of our framework's usefulness. 
	    Bottom-k methods have been used in numerous other applications such as
	    set similarity \cite{broder1997resemblance},
	    networking \cite{spring2000protocol}, 
	    time-decayed sampling \cite{cormode2008timeaggregates},
	    and
	    distributed sampling \cite{Kolonko2006WeightedReservoir,Efraimidis2006WeightedReservoir, tirthapura2011distributedreservoir}

\vspace{-0.1cm}
\section{Sampling design and estimation }
    We first review sampling without replacement and the use of fixed thresholds for drawing samples. 
    One key contribution of our paper is under what conditions can a random, adaptive threshold be treated as if it was a fixed threshold.
    
	The notion of weighted or unequal probability sampling is at the core of many sampling problems. Sampling more informative items with higher probability leads to lower variance estimates. 
	Choosing an appropriate measure of informativeness allows for accurate estimation for problems ranging from simple aggregates to complex machine learning models \cite{ting2018influencesampling}. %
	Unequal probability sampling can also arise in other situations, such as in stratified sampling where strata can have different sampling probabilities or in distributed sampling where nodes may make independent choices about the sampling rate.

	\subsection{Poisson sampling with thresholds}
	\change{The simple} sampling scheme where the inclusion decisions for each item is independent
	of all others \change{is of particular interest}. That is 
	\begin{align}
	    Z_i \sim Bernoulli(\pi_i)
	\end{align}
	\change{where $Z_i=1$ if item $x_i$ is included in the sample.}
	Such schemes are called Poisson Sampling designs. This is not to be confused with Poisson distributed random variables.
	In this case, the desired inclusion probability $\pi_i$  for each $x_i$ must be known in advance. 
	
	\change{A fixed threshold can be used to draw such a sample.}
	Associate an independent auxiliary variable $R_i$ for each item along with a fixed threshold $T_i$. The item $x_i$ is included if $R_i < T_i$. If $R_i$ is continuous and $F_i$ is its cumulative distribution function (CDF), then
	the probability $x_i$ is sampled is $p(R_i < T_i) = F_i(T_i)$. 
	Choosing a threshold such that $\pi_i = F_i(T_i)$ yields a sample from the desired sampling design. We call the variable $R_i$ the priority of $x_i$ and denote the inclusion of $x_i$ by $Z_i = 1(R_i < T_i)$.
    From this, it is easy to see how the threshold $T_i$ can be used to adjust the inclusion probabilities.

	\subsection{Sampling challenges and estimation}
	However, unequal probability sampling can lead to challenges in sampling and estimation.
	For example, under memory budget or sample size constraints, Poisson sampling can violate the constraints since there is some non-zero probability that all items are sampled. It is thus natural to consider samplers that draw fixed size samples. The natural extension of Poisson sampling to fixed sizes is 
     Conditional Poisson Sampling (CPS), which is obtained from a Poisson Sampling design conditional on the sample size being exactly $k$.
	The Conditional Poisson Sampling design has the attractive property of being the maximum entropy sampling procedure for a set of inclusion probabilities.
	However, there is no efficient algorithm known for drawing a CPS sample or for computing the inclusion probabilities \cite{tille2006sampling}.
	
	Estimating quantities of interest can be even more problematic. A sample is almost useless without good estimators. 
	In unequal probability sampling designs, an unbiased estimate of the population total $S$ is given by the Horvitz-Thompson (HT) estimator,
	\vspace{-0.15cm}
	\begin{align}
	\hat{S} = \sum_{i} x_i\frac{Z_i}{\pi_i}
	\tag{HT}
	\end{align}
	where $Z_i$ indicates if the item $x_i$ is included in the sample and $\pi_i = p(Z_i = 1)$ is the inclusion probability. 
	This also provides a solution to the subset sum problem \cite{duffield2007priority} by zeroing any value $x_i$ that is not in the desired subset.
	The fundamental problem for Conditional Poisson sampling and other sampling schemes with dependence between items' inclusion is that this dependence makes inclusion probabilities and good estimators difficult to derive. 
    
    \change{If the problem of efficiently sampling and computing inclusion probabilities can be overcome, then it is possible to search for an optimal sampling design given this estimator.}
	When the inclusion probabilities $\pi_i \propto x_i$, each term in the HT estimator is constant. Furthermore, if the sample size is fixed at $k$, the HT estimator itself is constant, and thus has minimal variance.
	A sample that draws elements with probability proportional to some size $x_i$ is called a probability proportion to size (PPS) sample.

	\subsection{Adaptive threshold sampling}
	\label{sec:adaptive threshold sampling}
	Designing good sampling procedures is further complicated by the fact that the desired inclusion probabilities are often not known in advance.
	They may depend on unknown properties of data stream such as length or the variance of the data. 
	
	Our solution to the challenge of adaptively choosing the desired probabilities is to replace the fixed threshold generating an independent Poisson sampling scheme with an adaptive threshold that can depend on the data.
	For example, adaptively choosing the threshold can ensure that the sample fits inside a memory budget regardless of the size of the stream.

	Mathematically, we define an adaptive 
	threshold $T_i = \tau_i(\mathbf{R} | \dcal)$ to be a function $\tau_i$ of the data $\dcal$ and priorities $\mathbf{R}$ that determine the sample. 
	Like Conditional Poisson Sampling, this dependence between the threshold and the priorities makes it difficult to compute the inclusion probability $p(Z_i = 1) = p(R_i < T_i)$ needed by the HT estimator and can make unbiased estimators difficult to generate. 
	For example, suppose the data consists of individuals' demographic information. In an extreme case, consider the threshold $T_i \defn \min \{R_j : \mathit{gender}_j = \mathrm{Female}\}$.  Such a sample is grossly biased as it would exclude all females. Generating an unbiased estimator of the population is impossible for this sampling scheme.
	
		Given the challenges introduced when samples are not drawn independently, we are interested in the following questions. 
	If the thresholds depend on the data, when can an unbiased or consistent estimator still be derived from the sample? Furthermore, when can the adaptive thresholds be treated as if they are fixed thresholds? An unbiased estimator for the fixed threshold would then automatically give an unbiased estimator for the adaptive threshold.

	\vspace{-0.1cm}
\subsection{Function classes}
\label{sec:function classes}
    Though the dependence between items makes computing arbitrary expectations difficult, we show \change{expectations are easy to compute} for restricted classes of functions. %
    We consider two function classes. 

    Our first goal is to examine when one can apply an unbiased estimator for a \emph{fixed} threshold sampler to get an unbiased estimate for an adaptive threshold sampler. 
    Given an adaptive thresholding scheme 
    , we seek to find the most general form of an estimator that allows the \change{such a} substitution.
    We first note that when sampling, any estimator is naturally restricted to be a function on the sample. \change{This general class of functions can be expressed as the polynomial 
    $\sigma_{\lambda \in \Lambda_0} f_\lambda(x_\lambda, T) \prod_{i \in \lambda} Z_i \prod_{i \not \in \lambda} (1-Z_i)$.}
    For the first class of functions, %
    \change{we consider a simpler class of polynomials of the form}
     \begin{align}
	\hat{\theta}(\mathbf{Z}, T) &= \sum_{\lambda \in \Lambda_0} \beta_\lambda(\mathbf{x}_\lambda, T_\lambda) \prod_{i\in \lambda} Z_i.
	\label{eqn: functions of sample}
    \end{align}
    Here, $\Lambda_0 \subset \pcal([n])$ where $\pcal([n])$ is the powerset of the indices $[n] \defn \{1, \ldots, n\}$. 
    A subscript of $\lambda$ selects the indices in $\lambda$.
    Since the priorites $R$ determine $Z$ and $T$ we also write the estimator as $\hat{\theta}(R,T)$ or $\hat{\theta}(R)$. 
    This form as a polynomial is particularly useful as the linearity of expectations allows each monomial term's expectation to be computed separately. 

    By further restricting the class of functions, we can capture a wider range of adaptive thresholding schemes where one has existing unbiased estimators.
    The second function class, the important set of pseudo-HT estimators, are of the form
     \begin{align}
    \hat{\theta}(\mathbf{R}) &= \sum_{\lambda \in \Lambda_0} h_\lambda(\mathbf{x}_\lambda) \prod_{i\in \lambda} \frac{\tilde{Z}^{\lambda}_i}{F_i(T^{\lambda}_i)}.
	\label{eqn: HT function class}
    \end{align}
    Rather than each item having a single threshold, an item's threshold can change depending on the term $\lambda$ in the sum.
    Since the thresholds can be different, the inclusion indications $\tilde{Z}^{\lambda}$ can be as well. 

    We note that the restriction to pseudo-HT estimators is mild. For any i.i.d. sample from a distribution $G$, \emph{any} estimable parameter of the distribution $G$ is equal to $\E h(X_1, \ldots, X_m)$ for some finite $m$ and symmetric function $h$ \cite{halmos1946theory}. This form, which is that of a U-statistic, is an even more restricted form \change{than} that of an HT-estimator.

\subsection{Threshold recalibration}
\label{sec: threshold recalibration}
    When using an adaptive threshold sampling rule, the main difficulty in computing expectations  is due to the interdependence of the threshold and priorities. 
    Our main idea is that, if we consider a monomial term $\Pi_{i \in \lambda} Z_i$, 
    the true adaptive threshold $T$ can be replaced with an alternative threshold $T^{\lambda}$ 
    that is independent of the priorities $R_\lambda$. \change{Here, superscript $\lambda$ is used to denote quantities whose values depend on $\lambda$, while subscripts select indices.}

    This can also be seen as creating an alternative threshold sampling procedure for every monomial term. Since each monomial corresponds to a subset $\lambda$ of all the items, 
    the alternative threshold can adapt to the data using only items not indexed by $\lambda$.
    As long as the alternative thresholds are not larger than the original ones, $\tilde{T}^\lambda_\lambda \leq T_\lambda$, the sample using the alternative can be computed from the sample using the original threshold. 
    We are particularly interested when the alternative and original thresholds are equal, $\tilde{T}^\lambda_\lambda = T_\lambda$. 

    These alternative thresholding rules are created in the following manner. 
    \change{Let $n$ be the possibly unknown number of items and $R_{-\lambda}$ denote the vector of priorities \emph{excluding} those indexed by $\lambda$.}
    Define the \emph{recalibrated thresholding rule} and threshold with respect to $\lambda$ by 
    \begin{align*}
        \tilde{\tau}^\lambda_i(R_{-\lambda}) &= 
        \inf_{r} \{ \tau_i(r): r_{-\lambda} = R_{-\lambda}\}
        \\
        \tilde{T}^\lambda &= \tilde{\tau}^\lambda(R_{-\lambda}) \leq T_\lambda
    \end{align*}
    In other words, given a set of priorities, \change{we find the smallest possible threshold with the given values of $R_{-\lambda}$ while ignoring the values of priorities $R_\lambda$.}
    When the thresholding rule $\tau$ is non-decreasing, the alternative threshold is obtained by setting every priority $R_i$ for all $i\in \lambda$ to the smallest possible value.
    These recalibrated thresholds yield a modified set of inclusion indicators $\tilde{Z}^\lambda_i = 1(R_i < T^{\lambda}_i)$.

    Although the true inclusion probability remains intractable to compute, computing a conditional inclusion probability is easy.
    \begin{lemma}
    The conditional inclusion probability given the recalibrated threshold is 
    \begin{align*}
        p\left(\prod_{i \in \lambda} \tilde{Z}_i = 1 \,|\, \tilde{T}^\lambda\right) &= \prod_{i \in \lambda} \change{p}(R_i < \tilde{T}^\lambda_i) = \prod_{i \in \lambda} F_i(\tilde{T}^{\lambda}_i), 
    \end{align*}
    \end{lemma}

    This provides the ability to estimate any statistic as long as the sample size is large enough.
    \begin{theorem}
    \label{thm:pseudo HT}
    Given a statistic $\theta = \sum_{\lambda \in \Lambda_0} h_\lambda(x_\lambda)$, the pseudo-HT estimator \change{below} using recalibrated inclusion indicators $\tilde{Z}$ and thresholds $\tilde{T}$ is an unbiased estimator for $\theta$.
    \begin{align}
        \hat{\theta}(\tilde{Z},\tilde{T}) &= \sum_{\lambda \in \Lambda_0} h_\lambda(x_\lambda) \prod \frac{\tilde{Z}^\lambda_i}{F_i(\tilde{T}^\lambda_i)}
    \end{align}
    \end{theorem}
    \begin{proof}
    Conditioning on $\tilde{T}^\lambda$ and applying the tower rule gives
    \begin{align*}
    \E \left(\prod \frac{\tilde{Z}^\lambda_i}{F_i(\tilde{T}^\lambda_i\change{)}}\right) &= \change{\E\,  \E \left(\prod \frac{\tilde{Z}^\lambda_i}{F_i(\tilde{T}^\lambda_i)} \bigg|  \tilde{T}^\lambda_i \right) = \E \left(  \frac{\E(\prod  \tilde{Z}^\lambda_i | \tilde{T}^\lambda_i) }{\prod F_i(\tilde{T}^\lambda_i\change{)}}  \right)} \\
    &= \change{
    \E \left( \frac{\prod F_i(\tilde{T}^\lambda_i)}{\prod F_i(\tilde{T}^\lambda_i)}  \right) }= 1.
    \end{align*}
    Thus, $\E\, \hat{\theta}(\tilde{Z},\tilde{T}) = \sum_{\lambda \in \Lambda_0} h_\lambda(x_\lambda) \cdot  1 = \theta$
    \end{proof}

    The lemma \change{trivially gives} an unbiased estimator for \change{subset} sums.
    \begin{corollary}
    \change{For any subset $\ccal \subset [n]$,} the following conditional HT-estimator based on the alternative thresholds is \change{an unbiased estimator for the subset sum $\sum_{i \in \ccal} x_i$}:
    \vspace{-0.05cm}
    \begin{align*}
	    \hat{\theta}_{HT}(R, \tilde{T}) &:= \sum_{i=\ccal} x_i\, \frac{\tilde{Z}^i_i}{F_i(\tilde{T}^i_i)}.
	\end{align*}
    \end{corollary}

    \vspace{-0.15cm}
	\subsubsection{Example: Bottom-k sketches and priority sampling}
	\label{sec:bottom-k example}
	Priority sampling and other bottom-k sampling procedures choose the threshold to be the $(k+1)^{th}$ smallest priority $R_{(k+1)}$. 
	Since the thresholding rule is non-decreasing, we can recalibrate the threshold for any item $x_i$ in the sample by changing the priority $R_i$ to $-\infty$. These priorities were already smaller than the $(k+1)^{th}$ smallest priority\change{, so} changing them to $-\infty$ does not affect the threshold, \change{and} $\tilde{T}^i_i = T_i$. Thus, the HT-estimator \change{is } unchanged \change{when the original threshold $T$ is substituted} with the recalibrated threshold \change{$\tilde{T}$}, and
	\begin{align}
	    \hat{\theta}_{HT}(R, T) &= \hat{\theta}_{HT}(R, \tilde{T})
	\end{align}
	is an unbiased estimator of the sum if $F_i(T_i) > 0$ %
	for all $i \in \ccal$.
	
	The main differences when deriving an unbiased estimator using threshold recalibration compared to \change{using} existing methods are that (1) threshold recalibration provides a constructive procedure for updating the thresholds while existing derivations must first propose a new thresholding rule that is then verified to match the existing one, and (2) we allow the alternative thresholds to differ for every monomial term.
	However, we are most interested in the case where the original thresholds and the recalibrated ones are the same.
	In this case, the original adaptive thresholds can be treated as fixed thresholds for the relevant class of functions.
	
\vspace{-0.1cm}
\subsection{Threshold substitutability}
    Suppose the recalibrated thresholds $\tilde{T}^\lambda$ are equal to the original ones whenever the subset $\lambda$ is in the sample. Formally, \change{for any subset $\lambda$,} $\change{Z}^{\lambda}_i = 1 \,\,\forall i \in \lambda \implies \tilde{T}^\lambda_\lambda = T_\lambda$.
    We call thresholds that satisfy this \emph{substitutable} thresholds. 
    If a threshold only satisfies this when $|\lambda| \leq d$ then we call it a $d$-substitutable threshold.
    Substitutable thresholds have several attractive properties.
     Most importantly, unbiased estimators derived under a fixed thresholding scheme are also unbiased under the true adaptive sampling scheme under modest regularity conditions.
     Thus, substitutable thresholds can be treated almost like fixed thresholds.

    \begin{theorem}[Threshold substitution]
    \label{thm:unbiased estimation}
    Let $T$ be a substitutable threshold.
    Suppose the estimator $\hat{\theta}$ in the form given in \ref{eqn: functions of sample}. 
    Then
    \begin{align}
        \hat{\theta}(R, T) &= \hat{\theta}(R, \tilde{T}).
    \end{align}
    This also holds if $\tau$ is $d$-substitutable and $\hat{\theta}$ is at most a $d$ degree polynomial in $Z$.
    \end{theorem}
    \begin{proof}
        Consider a term $\beta_\lambda(x_\lambda, T_\lambda) \prod_{i \in \lambda} Z_i$.
        If $\prod_{i \in \lambda} Z_i = 1$ then the \change{recalibrated threshold} $\tilde{T}^\lambda_\lambda = T_\lambda$ does not change. Thus, $\tilde{Z}^\lambda_\lambda =Z_\lambda$ as well, and
        $\beta_\lambda(x_\lambda, T_\lambda)\prod_{i \in \lambda} Z_i = \beta_\lambda(x_\lambda, \tilde{T}^\lambda_\lambda)\prod_{i \in \lambda} \tilde{Z}^\lambda_i$.
        Otherwise, if $\prod_{i \in \lambda} Z_i = 0$ then $\prod_{i \in \lambda} \tilde{Z}^\lambda_i = 0$ since recalibrated thresholds are always less than or equal to the original one.
    \end{proof}

    \begin{corollary}
    Let $T$ be a substitutable threshold.
    Suppose $\hat{\theta}(R, t)$ is in the form given in \ref{eqn: functions of sample}
    and is an unbiased estimator of $\theta$ for any value of $t \in Range(T)$.
    Then $\hat{\theta}(R, T)$ is an unbiased estimator of $\theta$.
    \end{corollary}
    \begin{proof}
    $\E \hat{\theta}(R, T) = \E \hat{\theta}(R, \tilde{T}) = \E(\hat{\theta}(R, \tilde{T}) | \tilde{T}) = \E(\theta | \tilde{T}) = \theta$ \change{where the first two equalities follows from Theorems \ref{thm:unbiased estimation} and \ref{thm:pseudo HT}}.
    \end{proof}

    The definition of a substitutable thresholding rule requires verifying that the original thresholds equal recalibrated thresholds with respect to every possible subset of indices. We now provide a simpler condition to verify substitutability that recalibrates with respect to singletons. 

\begin{theorem}[Substitutability from singletons]
\label{thm:singleton substitutability}
 Let $\tau$ be a non-decreasing adaptive thresholding rule generating the threshold $T$.
 If for any $i \in \{1, \ldots, n\}$, 
 $\tilde{T}^i_\lambda = T_\lambda$ whenever $\prod_{j \in \lambda} Z_j = 1$,
 then $T$ is a substitutable threshold.
\end{theorem}
\begin{proof}
Let $\tau(R) = T$ be the thresholding function for $T$.
We must verify $\tilde{T}^\lambda_\lambda = \change{T_\lambda}$ for all subsets $\lambda$ with non-zero probability of being selected.
Without loss of generality assume the subset to be verified is $\lambda = \{1, \ldots, k\}$.
If $\prod_{j \in \lambda} Z_j = 1$, then $R_j < T_j$ for all $j \in \lambda$. \change{The singleton substitutability assumption allows us to incrementally substitute one coordinate $R_j$ with any value $r_j < T_j$ without changing the threshold if $j \in \lambda$.}
Using induction, we have
$T = \tau(r_1, r_2, \ldots, r_k, R_{k+1}, \ldots, R_n)$ whenever $r_j < T_j$ for all $j \in \lambda$.
Since the recalibrated threshold $\tilde{T}^\lambda$ is simply the infimum over the coordinates indexed by $\lambda$, it follows that \change{$\tilde{T}^\lambda_\lambda = T_\lambda$}.
\end{proof}

\subsubsection{Variance of the HT estimator} 
The value of threshold substitution is that it makes it trivial to obtain unbiased estimators under adaptive threshold sampling. One can simply \change{reuse} an existing estimator for a simple, Poisson sampling design.
We illustrate the ease in estimating the variance of an HT estimator using our framework. In comparison, the priority sampling paper \cite{duffield2007priority} required a one and a half page derivation.

Section \ref{sec:bottom-k example} showed that the bottom-k and priority sampling threshold satisfies the conditions of Corollary \ref{thm:singleton substitutability}. Thus, it is substitutable.
The \change{well-known} variance of the HT estimator $\hat{\theta}_t$ under fixed threshold sampling with threshold $t$ and an unbiased estimator of this variance are given by
\begin{align*}
    Var(\hat{\theta}_t) &= \sum_{i \in \ccal} \left(\frac{1-F_i(t)}{F_i(t)} \right) x_i^2,
\quad \widehat{Var}(\hat{\theta}_t) \defn \sum_{i \in \ccal} \left(\frac{1-F_i(t)}{F_i(t)^2} \right) Z_i x_i^2 
\label{eqn:HT var estimate}
\end{align*}
The squared error $(\hat{\theta}(Z,T) - \theta)^2$ is in the function class given by equation \ref{eqn: functions of sample} \change{since the coefficient for any monomial $Z^\lambda$ depends only on $T_\lambda, x_\lambda$.
Hence,} the variance estimator for the HT estimator on fixed thresholds is also unbiased for the adaptive bottom-k threshold.
That is,
$Var(\hat{\theta}_T) = Var(\hat{\theta}_{\tilde{T}}) = \E Var(\hat{\theta}_{\tilde{T}} | \tilde{T}) = \E \widehat{Var}(\hat{\theta}_{\tilde{T}} | \tilde{T})$
\change{if $k \geq 2$}
due to the tower rule and unbiasedness of $\hat{\theta}$.

\subsubsection{Other statistics}
\change{
A slightly more complex case than a subset sum is Kendall's $\tau$ correlation. Given random $\mathbf{X}, \mathbf{Y} \in \mathbb{R}^n$,  Kendall's $\tau$ statistic is
\begin{align*}
    \tcal := \binom{n}{2}^{-1}\sum_{i < j} \sign(X_i - X_j) \sign(Y_i - Y_j).
\end{align*}
This is an alternative to the usual correlation 
$Cov(X, Y) / \sqrt{Var(X) Var(Y)}$. It can capture some non-linear dependencies and is robust to outliers.
It can be used to perform a hypothesis test whether two variables are dependent. %
If a threshold $T$ is 2-substitutable, then
an unbiased estimate of Kendall's $\tau$ is
\begin{align*}
    \widehat{\tcal} = {n \choose 2}^{-1}\sum_{i < j} \frac{C_{ij}}{F_i(T_i)F_j(T_j)}Z_i Z_j
\end{align*}
where $C_{ij} := \sign(X_i - X_j) \sign(Y_i - Y_j)$.
This is a form of HT-estimator, but the terms $Z_i Z_j$ in the estimator are correlated. This requires the more general form of the HT-estimator's variance:
\begin{align*}
    Var(\widehat{\tcal} | X, Y) &= {n \choose 2}^{-2} \left(\sum_{i \neq j} \frac{1-\pi_{ij}}{\pi_{ij}} C_{ij} + \sum_{i \neq j, k \neq \ell} \frac{\pi_{ijk\ell} -\pi_{ij}\pi_{k\ell}}{\pi_{ij}\pi_{k\ell}} C_{ij}C_{k\ell} \right)
\end{align*}
where $\pi_{ij} = P(Z_i = Z_j = 1)$ and
$\pi_{ijk\ell} = P(Z_i = Z_j = Z_k = Z_{\ell})$ 
are the pairwise and four-wise inclusion probabilities.}
\change{
Like above, %
we can apply Horvitz-Thompson under Poisson sampling to estimate this sum using only sampled items by replacing $C_{ij}$ with $C_{ij} Z_{i}Z_j / \pi_{ij}$ and $C_{ij}C_{k\ell}$ with $C_{ij}C_{k\ell} \frac{Z_{i}Z_jZ_kZ_\ell}{\pi_{ijk\ell}}$.
Given a substitutable threshold $T$, 
Theorem \ref{thm:unbiased estimation}
shows that we can use $\pi_{ij} = F_i(T_i)F_j(T_j)$
and $\pi_{ijk\ell} = \prod_{i' \in \{i,j,k,\ell\}} F_{i'}(T_{i'})$
Since $(\widehat{\tcal} - \tcal)^2$ is a $4^{th}$ degree polynomial in the function class given in equation \ref{eqn: functions of sample}, it is an unbiased estimator of the variance as long as the sample always contains at least $4$ items. 

Other statistics of interest include skew and kurtosis.
The skew is $\mu_3 / \sigma^{3/2}$ while the kurtosis is $\mu_4 / \sigma^4$.
Here, $\mu_k = \E(X - \E\,X)^k$ is the $k^{th}$ central moment and $\sigma^2$ is the variance. Like variance estimates, the sample central moments are biased estimators of a distribution's population moments. Our theory reuses existing unbiased estimators for the population moments $\mu_k$. For example, \cite{heffernan1997unbiased} provides an unbiased estimator for the $k^{th}$ central moment as a degree $k$ U-statistic. Section \ref{sec:function classes} showed that U-statistics have an unbiased HT-estimator. Hence,
plugging in adaptive thresholds into this estimator yields an unbiased estimate of the population moments whenever the threshold is $k$-substitutable. 

Each of these cases follow a simple paradigm. Adapt an existing estimate to use samples under a simple Poisson sampling design. Check a condition for the adaptive thresholds. If it holds, the estimator is unbiased when plugging in the adaptive thresholds.
}

\subsection{Sequential thresholding rules}
The last class of thresholds and functions  consists of thresholds whose values can be determined in a sequential manner. Although these thresholds may not be substitutable, we show they can yield unbiased HT-estimators. The following example \change{motivates this}.

Example (1-substitutable threshold):
Suppose a data stream is processed using a bottom-k sketch. Instead of storing only items in the final state of the bottom-k sketch, suppose an item is stored as long as it was in the bottom-k sketch at some point in the stream. This allows aggregates to be computed over time windows $[0,t]$ for any time $t$.
In this case, the threshold rule $\tau_i$ is a function of the priorities $R_1, \ldots, R_{\change{i-1}}$.
$\tau$ is trivially 1-substitutable \change{since it does not depend on $R_i$}. %
However, the threshold is not 2-substitutable. To see this, consider the state of the sketch after processing the entire stream, and let $x_j$ be the last item included in the sample.
If the data stream is sufficiently large, then the threshold $T_j$ must be equal to a priority $R_i$ of an item that appeared earlier in the stream and was included \change{in the} sample but was kicked out of the sketch before the end of the stream. If that earlier priority $R_i$ changes, then the later threshold $T_j$ also changes. 
Thus, while 1-substitutability allows us to use the HT estimator for sums, it does not allow us to compute \change{unbiased estimates of} variances.

However, despite being non-substitutable, we show this threshold can still be treated like a fixed threshold for pseudo-HT estimators. We show that if there is an ordering of the data such that the 
thresholding choices are made sequentially, then a pseudo-HT estimator is unbiased.
That is, we must show $\E \prod_{i \in \lambda} Z_i / F_i(T_i) = 1$ for any $\lambda \subset \Lambda_0$ where 
$p(\prod_{i \in \lambda} Z_i = 1) > 0$.

\begin{theorem}
Given a permutation $\rho_1, \ldots, \rho_n$ of $[n]$, define \change{the future samples at time $j$ to be} $S^j\change{(\rho)} = \{\rho_k: Z_{\rho_k} = 1,\, k \change{\geq} j\}$.
If \change{there exists a permutation $\rho$ such that} the recalibrated thresholds $\tilde{T}^{S^j\change{(\rho)}}_k = T_k$ for all \change{$k \leq j$} 
with $Z_{\rho_k} = 1$, then 
\begin{align}
    \E \prod_{i \in \lambda} \frac{Z_i}{F_i(T_i)} = 1
\end{align}
for any $\lambda$ such that $p( \prod_{i \in \lambda} Z_i = 1) > 0$.
\end{theorem}
\begin{proof}
This is a straightforward application of Fubini's theorem. 
\begin{align*}
\E \prod_{i \in \lambda} \frac{Z_i}{F_i(T_i)}
&= \E, \E\left(\prod_{i \in \lambda} \frac{Z_i}{F_i(T_i)} \bigg| R_i, i \not \in \lambda \right) 
    = \E \int \prod_{i \in \lambda} \left( \frac{Z_i}{F_i(T_i)} dF_{i}(R_{i})\right) \\
    & = \E \int \left(\prod_{i \in \lambda \backslash \{\rho_k\}} \frac{Z_i}{F_i(T_i)} dF_{i}(R_{i})\right) \frac{F_{\rho_k}(T_{\rho_k})}{F_{\rho_k}(T_{\rho_k})} \\
    &\quad \cdots = \E 1 = 1
\end{align*}
\change{where $k$ denotes the largest index such that $\rho_k \in \lambda$. The terms $Z_i / F_i(T_i)$ for $i \in \lambda \backslash \{\rho_k\}$ are unaffected by integration over $R_{\rho_k}$ since 
$\rho_k$ is the index for a future sample for $i$,
and the recalibration condition ensures that modifying priority $R_{\rho_k}$ does not affect the threshold $T_i$ as long as $R_{\rho_k} < T_{\rho_k}$.}
\end{proof}

\change{In one special case where the permutation puts the priorities in sorted order, a stronger full substitutability result is obtained.}

\begin{theorem}
\label{thm:filtration threshold}
Consider the sequence $R_{\rho_1} > R_{\rho_2} > \cdots > R_{\rho_n}$.
If $M$ is a stopping time with respect to the filtration defined by this sequence, then
the rule $\tau(\mathbf{R}) = R_{\rho_M}$ is a substitutable threshold.
\end{theorem}
\change{\begin{proof}
The items $\lambda = \{\rho_{M+1}, \ldots, \rho_{n}\}$ are in the sample.
By the definition of a stopping time, the threshold $R_{\rho_M}$ is a function of $R_{\rho_1}, \ldots, R_{\rho_M}$.
Furthermore, it is non-increasing.
Applying threshold recalibration and replacing $R_j$ with $-\infty$ for all $j > M$ does not change the ordering of the first $M$ priorities. Thus, the threshold does not change, and the threshold is substitutable.
\end{proof}
}

\subsection{Building and composing thresholds}
\label{sec:building}
Thus far, we have described methods for taking an existing thresholding rule and modifying it to allow for unbiased estimation.
We now examine how thresholds can be composed and samples can be merged. A simplified interpretation of our result states that the maximum of 1-substitutable thresholding rules yields another 1-substitutable rule, \change{and} the minimum of fully or d-substitutable rules yields another fully or d-substitutable rule respectively. \change{Composing $d$-substitutable or fully substitutable thresholding rules preserves $d$-substitutability or full substitutability.}

\begin{theorem}
Let $T^1, T^2$ be $1$-substitutable thresholds on a data set \change{$\dcal$ with priorities $\mathbf{R}$}. \change{Let the items sampled according to $T^j$ be denoted by $\dcal^j$
and their indices by $\ical^j$}.
Consider 
 a thresholding rule $\tau'$ where $\tau'_i$ is a function of $\dcal^1, \dcal^2$ and $T^1_i, T^2_i$.
\change{Let} $T'$ \change{be a threshold such that $T'_i = \tau'_i$ for $i \in \ical^1 \cup \ical^2$
and $T'_i \leq \max\{T^1_i, T^2_i\}$ for all $i$.}
If $\tau'$ is a 1-substitutable \change{thresholding} rule for the data set \change{$\dcal^1 \cup \dcal^2$}
then $T'$ is 1-substitutable threshold on the original dataset.
Likewise, if $T^1, T^2$ are substitutable, $\tau'$ is substitutable \change{for $\dcal^1 \cup \dcal^2$,
and $T'_i \leq \min\{T^1_i, T^2_i\}$ for all $i$},
then $T'$ is substitutable.
\end{theorem}
\begin{proof}
For 1-substitutability, note that changing the priority $R_i$ for any $i \in \ical^1 \cup \ical^2$ does not change $T_i^1, T_i^2$. Hence, it also cannot change $T'_i$. Similarly, when $T^1, T^2$ are substitutable, \change{one can recalibrate priorities for items in $\dcal^j$ without affecting $T^j$. Hence, recalibrating priorities for items in $\dcal^1 \cap \dcal^2$ affect neither $T^1$ nor $T^2$. These are precisely the set of items with priority $R_i \leq \min\{T_i^1, T_i^2\}$.}
\end{proof}

\subsection{Priority-threshold duality}
Another useful property of adaptive threshold sampling is that adjusting priorities is equivalent to adjusting thresholds. An item with a priority distribution $F_i$ and per item threshold $T_i$ is included if $R_i = F_i^{-1}(U_i) < T_i$.
Equivalently, it is included if a random uniform random variable $U_i < F_i(T_i)$ 
is less than the pseudo-inclusion probability.
\change{This can be used, for example, when}
the importance of an item and its priority distribution can change over time. 

For instance, in time-decayed sampling \cite{cormode2008timeaggregates} with exponential decay, the weight of an item $w_i(t) = \overline{w}_i \exp(-t)$ decreases exponentially with time $t$ after the item appear at time $t_i^0$. 
Thus, one can build a sampling method which uses adaptively chooses a threshold $T(t)$ given time varying weights $w_i(t)$. An item is included in the sample at time $t$ if its priority $R_i(t) = U_i / w_i(t) < T(t)$
However, it is inconvenient in practice to change the weight of existing points. Changing the threshold to increase exponentially instead, allows the priorities to remain fixed. An item is included if $\overline{R}_i = U_i / \overline{w}_i < \exp(t) T(t)$. 

	\section{Applications and Sampling designs}
	Although the definitions and theorems in section \ref{sec:adaptive threshold sampling} are subtle and abstract, we now show that they are powerful, allowing us to easily generate sampling schemes that solve novel problems$^\dagger$, improve existing sketches$^\ast$, and unify the theory for multiple sampling methods$^\ddagger$. We label the sections with the respective symbol $^\dagger$, $^\ast$, or $^\ddagger$ to note the contribution our paper makes.
	
	\subsection{Variable item sizes$^\dagger$}
	Bottom-k sketches ensure the sample size is always $k$. However, the memory usage of the sample can vary if items \change{have} different sizes. To guarantee a sample fits within \change{a given memory budget $B$}, the parameter $k$ must \change{be} set conservatively to $B/L_{max}$ where $L_{max}$ is the size of the largest item. This is highly inefficient if the largest item is much bigger than the average item.
	Another thresholding rule $\tau$ simply takes as many items as possible that fit within the memory budget.
	If priorities are sorted in ascending order,
	the threshold is the priority of the first item which causes the budget to be exceeded. Like a bottom-k sketch, the values of the smaller priorities are irrelevant and can be set to 0, so the threshold is substitutable. Thus, if the budget $B \geq L_{max}$ so every item has a non-zero \change{chance} of being selected, the usual HT estimator provides estimates of subset sums, and if $B \geq 2\,L_{max}$ the usual HT variance estimator provides unbiased estimates of its variance.
	
	For example, 
 items in the 2020 Kaggle data science survey can vary in size since the survey contains \change{both long,} text free responses and cases where the respondent does not finish the survey. As a string, the maximum length of an item is 5113 characters while the average length is 1265. A bottom-k sample that is guaranteed to fit within a budget constraint is expected to be $1/4^{th}$ the size of an adaptive threshold sample that utilizes the entire budget.

	\vspace{-0.1cm}
	\subsection{Sliding windows$^\ast$}
	Oftentimes, only recent items in a data stream are of interest. 
	A sliding window sampler draws a uniform sample from points that arrive in the time interval $(t-\Delta, t]$ where $t$ is the current time and $\Delta$ is the length of the time window. When the arrival rate of items changes over time, it can be impossible to draw a fixed size sample in bounded space \cite{gemulla2008sampling, braverman2009optimal}.
	The state-of-the-art for drawing uniform samples from sliding windows in bounded space is given by Gemulla and Lehner (G\&L) \cite{gemulla2008sampling}. We show this is an instance of adaptive threshold sampling, but a highly inefficient one. Our framework immediately yields improved thresholds that double the number of usable points with zero modifications to the sketch. 
	
	 At time $t$, the G\&L scheme consists of one set of expired samples $X(t)$ that occur in the time window $(t-2\Delta, t-\Delta]$ and another set of current examples $C(t)$ in the window $(t-\Delta, t]$.  
    Although G\&L do not describe their procedure as a thresholding scheme, we describe it as a two stage thresholding scheme, one which samples non-uniformly to build candidate points and one which provides the final uniform sample.
    
    The initial threshold $T_n(t_n)$ for an item $x_n$ at time $t_n$ is $1$ if there are fewer than $k$ current examples.
    Otherwise, $T_n(t_n)$ is the $k^{th}$ smallest priority  of the current sample $C^{\change{-}}(t_n)$ just before time $t_n$ and the new priority $R_n$. \change{We note that this can be larger than the $k^{th}$ smallest priority for all items in the current time window. Some items not in the current sample may have a smaller threshold and have already been discarded.}
    If there are ever more than $k$ current examples, the largest priority item is discarded by adjusting the threshold of all the current examples,
    $T_i(t_n) = \min\{T_i^-(t_n), T_n(t_n)\}$.
    An example that falls out of the current window is moved from the current to expired examples, and any expired item that is two window lengths or more from the current time is discarded. This threshold determines whether or not an item is stored but does not ensure that it is a uniform draw from a sliding window. 

The G\&L scheme assigns a final threshold $T_{GL}$ equal to the $k^{th}$ smallest priority of the combined current and expired examples. This is guaranteed to return a uniform sample from the current time window, although the \change{sample} size is not fixed. In this case, the item corresponding to the threshold can be included in the sample due to symmetry. %

This bottom-k threshold, however, results in an inefficient threshold that discards half of the useful points. \change{To see this, suppose that items arrive at a uniform rate of 1000 per second and are assigned $\mathit{Uniform(0,1)}$ priorities. Given a 100 second time window and a maximum memory budget of $1000$ samples in the current window, the initial threshold for each item must be roughly $T_i \approx 1\%$ to satisfy the budget. However, G\&L takes the bottom-1000 threshold of items in both the current \emph{and expired} windows, which contains 200,000 items. This gives a threshold of roughly $0.5\%$. Since all thresholds are $T_i \approx 1\%$ and we simply take the min of thresholds in the current window, we will return a threshold near the ideal threshold of $1\%$.}

Our adaptive thresholding framework immediately provides a much improved threshold. 
We note that 
the thresholding rule consists of two parts, a sequential sampling rule, and a sequence of minimum operations on the thresholds. 
The sequential rule generates 1-substitutable thresholds. 
and taking the min of thresholds preserves 1-substitutability. 
Thus, taking another min of all the thresholds in the current examples yields another 1-substitutable threshold. 
That is $T_{\mathit{improved}}(t) = \min_{i \in C(t)} T_i(t_n)$. 
\change{This threshold $T_{\mathit{improved}}(t)$ is constant over the
current time window $(t-\delta, t]$. Thus, 1-substitutability implies full substitutability by theorem \ref{thm:singleton substitutability}, and the threshold yields a uniform sample.}
\change{Since} the per-item threshold of the current examples can be calculated from the expired examples and earlier current examples, computing the improved threshold requires no additional storage, and the adaptive sampling framework provides the improvement for free.

\begin{figure}
\centering
	\begin{tabular}{cc}
		\includegraphics[width=.8\textwidth, height=.6\textwidth]{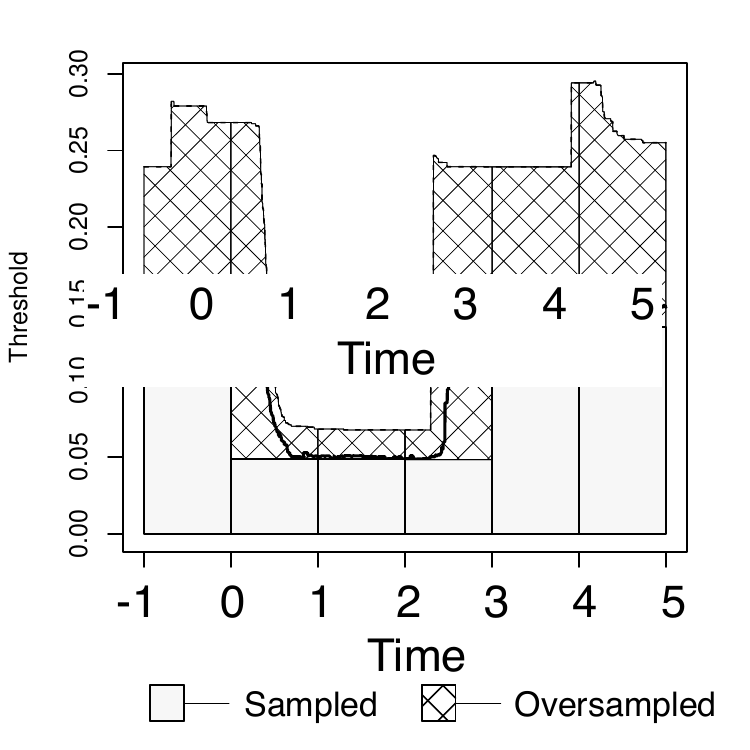} 
	\end{tabular}
	\caption{The top line shows the true marginal sampling probabilities which our method recovers. The middle line shows the conservative estimate used by the G\&L scheme, and the  boxes at the bottom show the thresholds used by a set of sliding windows. The hatched area above each box shows the amount of oversampling used to draw that sliding window sample. Note that even if the true thresholds were used, the scheme must still significantly oversample when the item arrival rate changes. }
	\label{fig:oversampling}
\end{figure}

\begin{figure}
	\includegraphics[width=.8\textwidth, height=.7\textwidth]{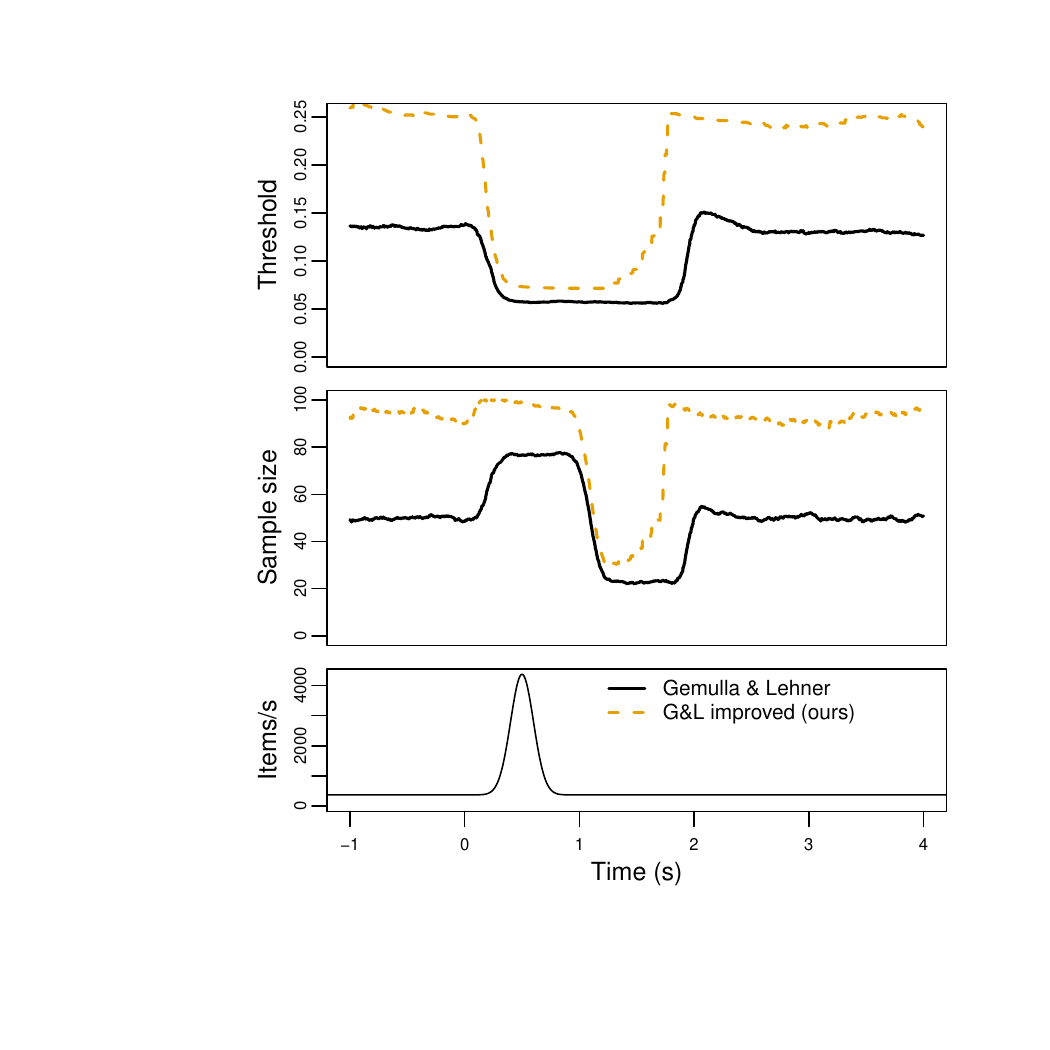} 
	\caption{Our improvement on G\&L draws more samples (middle) while recovering more quickly from a spike in the item arrival rate (bottom). This is because G\&L gives an underestimate of the threshold (top).  }
	\label{fig:gemulla}
\end{figure}

Figure \ref{fig:oversampling} shows the evolution of the per item \change{thresholds} $T_i(t_i)$ over time
and the much smaller threshold used by G\&L to construct a sliding window sample. 
Figure \ref{fig:gemulla} shows the behavior when there is a spike in the item arrival rate. 
Not only does our adaptive threshold sampling framework yield nearly twice as many samples when the item arrival rate is steady, it recovers from the spike faster.

	\subsection{Adaptive sampling for top-k and disaggregated subset sums$^\dagger$}
	Adaptive thresholding can also be used to modify a frequent item sketch into a top-k sketch that also supports further aggregations. 
	Given a parameter $m$, a frequent item sketch returns all items where the proportion of times each appears is $> 1/m$. This can be done with $O(m)$ space
	using the Misra-Gries \cite{misra1982frequent} or equivalent Space-saving sketch \cite{metwally2005spacesaving}. The top-k problem requires returning the top $k$ items by frequency. \change{Unlike the frequent item problem,} there is no guarantee on the minimum proportion of times these top-k items appear. This makes the top-k problem a more challenging variation of the frequent item problem.
	\change{To use} frequent item sketches for the top-k problem, the appropriate size parameter $m$ \change{must be known} in advance. 
	It can also be useful for the sketch to support further aggregations. For example, one may wish to analyze web page impressions and find the frequently viewed pages. One may then wish to further aggregrate pages by topic. This disaggregated subset sum problem is addressed by \cite{ting2018spacesaving, li2020wavingsketch}.

    This problem can be seen as an adaptive sampling procedure which learns to downsample infrequent items, leaving the frequent ones. By virtue of being an adaptive threshold sampling procedure, it automatically supports unbiased estimation of counts and further aggregations using the HT-estimator.
    This leads to the following adaptive threshold sampling procedure. 
    For each point $x_t$ in a data stream, assign a $Uniform(0,1)$ priority $R_t$.
    Maintain a variable length list where each entry consists of an item $x_i$, its priority $R_i$, a threshold $T_i$, and a count $v_i$ of the times it \change{appeared} after entering into the sample. An unbiased estimate of the count of an item is $\hat{c}_i = 1/T_i + v_i$. \change{This is the Horvitz-Thompson estimate where one appearance of the item has pseudo-inclusion probability $T_i$ while the other $v_i$ appearances have probability 1.}
    We define the adaptive threshold $T(t)$ at time $t$ to be the smallest priority such that at least $k$ items in the sample have estimated count $\hat{c}_i > 1/T(t)$.
    This splits the items in the sample into infrequent items with $\hat{c}_i \leq 1/T(t)$
    and $k$ frequent items with $\hat{c}_i > 1/T(t)$. Whenever the adaptive threshold $T(t)$ is updated, only infrequent items are updated. Those with priority $R_i \geq T(t)$ are discarded. All others update their threshold to $T(t)$ and their counter  to $v_i = 0$. 
    
    This procedure can be seen as a thresholding based variation of Unbiased Space-Saving \cite{ting2018spacesaving}. In both, 
    infrequent items form a random sample of the items not assigned to a frequent item counter. Items start as infrequent items and maintain a count of the number of times each occurred after entering the sample.
    It is easy to see that this thresholding rule is substitutable. For any subset of items in the sample, changing their priorities to $0$ has no effect on the sample or thresholds.
    Thus, like Unbiased Space-Saving, it can be used for unbiased estimation for the disaggregated subset sum problem.
    
    We compare our adaptive procedure with the FrequentItems sketch in Apache Datasketches \cite{apachedatasketches, anderson2017high}. This FrequentItems sketch is a variation of the Misra-Gries sketch \cite{misra1982frequent, metwally2005spacesaving} that allows for faster updates. For each sketch, we query for the top-k items in a stream with $k=10$ and record the number of errors in the result.
    Since we wish to compare performance as the distribution of the heavy hitters changes, we use a synthetic Pitman-Yor$(1,\beta)$ preferential attachment process that is able to generate both light tailed and heavy tailed behavior. It is commonly used in Bayesian cluster models. Larger values of $\beta \in [0,1)$ result in heavier tails. More precisely,
    in the Pitman-Yor process, the $t^{th}$ item in the stream is a new item with probability $(1+\beta C_t)/t$ where \change{$C_t$} is the number of unique items seen already.
    Otherwise, it is equal to the $j^{th}$ unique item with probability $(n_{tj} - \beta)/t$ where $n_{tj}$ is the number of times unique item $j$ has been seen amongst the first $t-1$ items. 

        \begin{figure}
        \centering
    \includegraphics[width=0.8\textwidth] {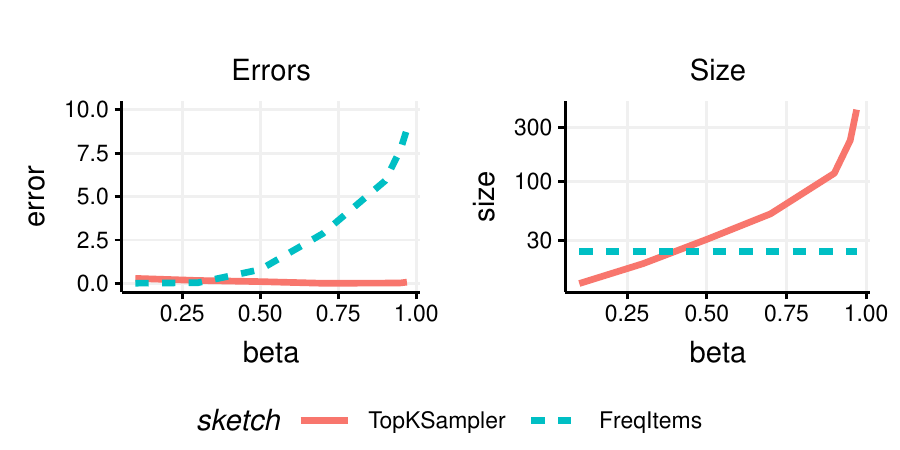}
        \caption{Comparison of the fixed size FrequentItems sketch and our adaptive Top-K sampler as the distribution of frequent items changes. 
        Smaller $\beta$ values have a few dominant heavy hitters while larges values yield more evenly distributed frequencies.
        Left: The average number of incorrect items in the top-k items returned by each sketch. Right: The size of the sketch in number of items.}
        \label{fig:topk}
    \end{figure}

    Figure \ref{fig:topk} shows that our procedure can accurately capture the top-k items without prior knowledge of the distribution. It does so by appropriately adjusting the size of the sample so that it captures the top-k items with high probability. For distributions where the the frequent items are well-separated from the remaining items, our adaptive sampler requires even less space than the FrequentItems sketch. 
    For distributions where frequent items are not well separated from infrequent items, the FrequentItems sketch performs poorly. In contrast, our adaptive procedure captures the frequent items
    by adjusting its size to the data. For FrequentItems, we take the size to be $0.75$ times the size of the allocated hash table.

	\subsection{Distinct counting for weighted samples$^\ddagger$}
	The subset sum and distinct count problems are often treated as separate problems. However, they can be addressed using a single weighted, coordinated priority sample. \change{For example, in a freemium model, one may wish to sample paying users with probability proportional to spend but may still wish to estimate the total population size of both paying and non-paying users in a demographic subgroup.}
	\change{Draw an adaptive threshold sample with } substitutable threshold $\mathbf{T}$.
	\change{Estimate the distinct count of items by } $\hat{N} = \sum_{i} Z_i / F_i(w_i T_i)$.
	\change{For any subset $\acal$, estimate the subset sum for the subset of indices $\acal$ by} $\hat{S}(\acal) = \sum_{i \in \acal} w_i Z_i / F_i(w_i T_i)$. 
    This extends the Theta sketch framework \cite{dasgupta2016thetasketch} to non-uniform priorities and weighted samples and allows for per item thresholds.
    
    \subsection{Improved merges for distinct counting$^\ddagger$}
\begin{figure}
        \centering
 \includegraphics[width=3.4in] {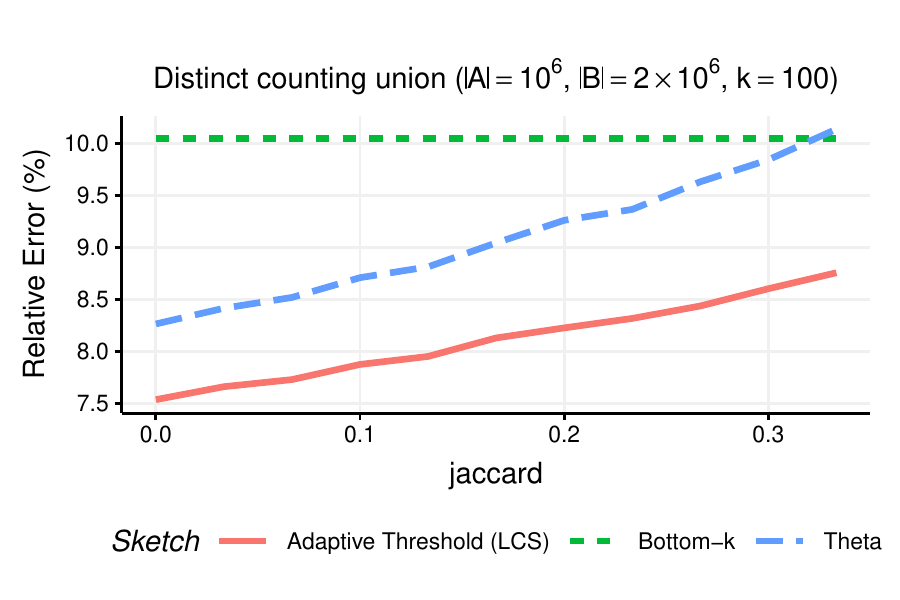}
        \caption{Relative error $SD(\hat{N} - N) / N$ as a function of the Jaccard similarity. Our adaptive thresholding procedure is the same as LCS in this case and improves upon the basic bottom-k sketch and the Theta sketch except when $A \subset B$.}
        \label{fig:distinct}
    \end{figure}

    The framework also can be used to improve merge procedures for distinct counting sketches based on coordinated samples, such as the MinCount/bottom-k sketch \cite{giroire2009order, beyer2009distinct, dasgupta2016thetasketch}. Given two coordinated samples with 1-substitutable thresholds $T^A$ and $T^B$ for sets $A$ and $B$, one simply needs to produce a new 1-substitutable threshold with $T'_i \leq  \max \{T^A_i, T^B_i\}$ to produce another distinct counting sketch. This generalizes the LCS sketch of \cite{cohen2009leveraging} which specifically takes the max of bottom-k thresholds. Our use of arbitrary 1-substitutable thresholds also allows merges to be chained together.
	Figure \ref{fig:distinct} illustrates the improvement when taking the union of sets $A$ and $B$ of size $|A| = 10^6$, $|B| = 2 \times 10^6$, and varying Jaccard similarity. \change{Another scenario where improved merges can help is when one set dominates the others in size. For example, if one set has size $|A_0| = 10^6$, and there are $10^6$ sets with $100$ distinct items, then Theta sketches of size 100
	will result in a threshold of $\approx 100 / 10^6 = 0.01$ after merging. It will estimate the population of $101 \times 10^6$ with error $\pm 1\%$.
	On the other hand, only the large sketch contributes to the error in our case. This error is $\approx 10^6 \times 1\%$, which is $100\times$ better than that of the Theta sketch.
	}

	\subsection{Frequent items for distinct counting$^\dagger$}
	A common database aggregation computes a distinct count of \change{items} grouped by some attributes, for example, the number of distinct users that saw an ad grouped by time and demographic variables. Although each distinct counting sketch is small, the group by operation can create tens of millions of sketches, resulting in a large memory footprint. Oftentimes, many of these groups contain only a few items.
	Previous approaches have used counter sharing \cite{xiao2015vhll, ting2019distinctbillions} and sparse representations for small counters \cite{Heule2013}. We propose a novel approach use subsampling.
	
	At any point in time, rather than maintaining a bottom-k sketch \change{of item hashes} for every group, 
	\change{maintain $m$ bottom-k sketches, each associated to a single group, and one general pool of samples. 
	For these $m$ groups, denote the \change{bottom-k} threshold for group $g$ by $T_g$. For the general pool, we use the threshold $T_{max} = \max_g T_g$.
	
	For a new item $x$ in group $g$, if $g$ corresponds to one of the $m$ bottom-k sketches, then add $x$ to the sketch for $g$.  Otherwise, add the item and its group to the general pool if its priority is below $T_{max}$. If there are more than $k$ items from the same group $g$ in the general pool, then let $g'$ be the group with the largest threshold amongst the $m$ and create a new bottom-k sketch for $g$ using items from the general pool. Replace $g'$'s counter with the bottom-k sketch for $g$, and move the items in the sketch for $g'$ into the general pool. In both cases, if  the threshold $T_{max}$ for the general pool is reduced, discard any items with priority above the new threshold.}
	In effect, we adjust the sampling rate to be the appropriate sampling rate for the top $m$ groups.
	Equivalently, the tolerated error for a small group is raised from being a percentage of the small group's size to a percentage of the heavy hitters' group sizes.
	
	While using a bottom-k sketch for every group forces the number of sketches to grow linearly with the number of groups, 
	many small groups will not have any sampled items \change{in this sketch}.  This alleviates the problem of having too many counters. %
	
    \subsection{Multi-Stratified sampling$^\dagger$}
	Suppose a data set of users can be stratified in two ways, by country or by age. We wish to draw a single sample which is both a stratified sample by country and a stratified sample by age and fits within a budget of $B$ items. We first show how to generate a stratified sample, and then how to control the budget.
	
	First, maintain a bottom-k 
	threshold $\tau^{(country)}_c$ for each country $c$ and $\tau^{(age)}_a$ for each age $a$.
	A user $x_i$ with country $c_i$ and age $a_i$ has a per item threshold of $\tau'(\mathbf{R}) = \max\{\tau^{(country)}_{c_i}(\mathbf{R}), \tau^{(age)}_{a_i}(\mathbf{R}) \}$.
	\change{Taking the maximum of substitutable thresholds yields a 1-substitutable threshold.
	Theorem \ref{thm:singleton substitutability} shows  the threshold is fully substitutable.}

	This sample has variable size $m \in [k \max \{n_c, n_a\},  k (n_c + n_a)]$
	where $n_c$ and $n_a$ are the number of distinct countries and ages.
	We wish to ensure the sample contains exactly $B$ items.
	We modify the stopping rule for the thresholds so that $k$, the number of items per stratum, is dynamically selected. For a set of thresholds, choose a stratum with the most number of elements below its threshold. Decrement its threshold to the next smaller priority. Since an item belongs to more than one stratum, one for country and one for age, this may not decrease the total sample size.
	Continue decreasing the thresholds until the desired sample size is reached. \change{A nearly identical argument to the bottom-k case shows that this threshold is also substitutable.}

	\subsection{Multi-objective samples$^\ast$}
	When using samples, the importance of an item may depend on the query. For example, 
	an analyst may be interested in either profit or revenue. Queries on profit ideally utilize a weighted sample with weight proportional to each item's profit. Similarly, queries on revenue ideally weight by revenue. 
	The existing approach by Cohen \cite{cohen2015multi} combines two coordinated bottom-k sketches,
	one \change{with weights optimized} for profit and another \change{optimized} for revenue, to obtain a sketch that has size $\leq 2k$.
	This ensures the combined sketch never does worse than an individual sketch.
	However, as more objectives are added, each objective's sketch must be made smaller.
	Given a budget constraint $B$ and $c$ different objectives, each objective can only be allocated a sketch with space $B/c$. When sketches have high overlap, the size of the combined sketch can be much less than the budget. For example, if every objective assigns \change{highly correlated weights to} each item, then the priorities \change{are highly correlated as well}. \change{Hence}, sketches for every objective are \change{nearly} the same and \change{approximately} $B/c$ of the budget $B$ is used after combining the sketches. \change{In the case where weights are scalar multiples of each other, exactly $B/c$ of the budget is used up to rounding}.

	\subsection{Variance sized samples$^\dagger$}
	\label{sec:variance sized samples}
	Priority sampling provides a relative error guarantee on the sum provided the weights are proportional to the values in the sum. It guarantees the variance of the error $\epsilon$ is bounded by $V(\epsilon) \leq S^2/(k-1)$ where $S$ is the true sum and $k$ is the sample size \cite{szegedy2006dlt}.
	However, one may wish to have a guarantee on the absolute error $V(\epsilon) \leq \delta^2$\change{,} or 
	the weights may not be proportional to the weights.

	We instead set a stopping time which stops at the first threshold $T$ where the estimated variance matches the desired error,
	$V(\hat{S}_T) \geq \delta^2$.
	Here, $\hat{S}_t$ is the HT estimator with a fixed threshold $t$.
	The unbiased estimate of the variance of the HT estimator $\hat{V}(\hat{S}_t) = \sum_i 1(R_i < t) x_i^2 \frac{1-wt}{wt} 1(wt < 1)$ is  
	discontinuous only at jumps when the threshold is equal to some priority. At all other threshold values, it increases continuously as $t$ decreases. Thus, $\E V(\hat{S}_T) = \delta^2$.

	We note that while processing a stream or data file, it is typically impossible to verify that a threshold is a stopping time with just the information in the sample. 
	The stopping time may be a larger threshold that includes additional points that are not in the sample. 
	However, since the true variance of an estimator is strictly increasing as the threshold decreases, it is reasonable to expect that all thresholds where the estimated variance equals the desired variance are close together. 
	By slightly oversampling, one is likely able to recover the true stopping time.
	
	\subsection{Early stopping in AQP$^\dagger$}
	Rather than explicitly constructing samples, another way to use adaptive threshold sampling is to store all of the data but sort items by their priorities. Given a user specified standard error $\delta$, an approximate query processing \change{(AQP)} system can provide an answer by using the variance sized sampling scheme given above to determine how many items need to be read and \change{to} stop processing once it has read enough.

	This can also be combined with other methods to define a physical layout of the data that is appropriate for sampling. For example, one can combine it with multi-objective sampling to get a layout that is useful for multiple queries. Suppose both revenue and quantity are metrics of interest in a dataset. Denote their values for row $i$ by $s_i$ and $q_i$ respectively and their priorities by $S_i = U_i / s_i$ and $Q_i = U_i / q_i$. 
    One can generate a file block consisting of a bottom-k sample ordered by $S_i$ and a bottom-k sample of the remaining items ordered by $Q_i$. Repeating this procedure on the remaining items generates a physical layout of the data which only needs to read $m$ blocks to get a weighted sample of size at least $mk$. 

	\section{Asymptotic behavior}
	\label{sec:asymptotics}
	The framework we have described thus far covers unbiased estimation. This excludes a number of important estimators that are consistent but not unbiased, such as maximum likelihood estimators, quantile estimators, and more generally, any estimator which is the maximizer of an objective function.
	We extend our framework to justify \change{using} such estimators as well as \change{using} some heuristic thresholding rules in asymptotic settings. We also show that in asymptotic settings where the inclusion probabilities go to 0, all priority distributions are asymptotically equivalent to \change{appropriately} weighted uniform distributions used in priority sampling.

	While the practical implication of this section are significant, it is largely theoretical and requires knowledge of empirical process theory. 
	For this reason, we first provide a brief overview of the main ideas. \change{It} requires minimal prerequisite knowledge and states our main results in a easily understandable way at the cost of rigor. 
	We then provide an overview for the theory we develop and techniques we use before stating and proving our results.
	
	\subsection{Basic Overview}
	The main idea is to consider adaptive thresholds that \change{either} converge to a fixed threshold with high probability \change{or are close to it. We show in} what ways they can be treated as if they were actually fixed thresholds.
	For example, if a thresholding rule generates a single threshold $T$ that is used for all points, 
	the threshold $T$ approximates a fixed threshold $t_0$ if 
	$T \in (t_0-\epsilon, t_0+\epsilon)$ with high probability (w.h.p.) for some small $\epsilon \geq 0$. 
	If $\hat{\theta}_t$ is an estimator for $\theta$ for fixed thresholds $t$. When applied to an adaptive threshold, the error of $\hat{\theta}_T$ is bounded by the worst case error
	$|\hat{\theta}_T - \theta| \leq \sup_{|h| < \epsilon} |\hat{\theta}_{t+h} - \theta|$
	for estimators on that range of thresholds.
	
	The difficulty \change{is} that we must show that the estimate at all perturbed thresholds $\hat{\theta}_{t+h}$ are close to $\hat{\theta}_t$. This requires a notion of continuity which cannot be obtained from a pointwise variance calculation at a specific threshold. \change{However, } empirical process theory allows us to do just that.
    
    We apply it to analyze the class of M-estimators, that is estimators that maximize an objective function formed by the sum over independent random variables. They are of the form 
	\begin{align}
	\label{eqn:M-estimator}
	J_n(\theta) &= \E_n f_\theta(X) = \sum_{i=1}^n f_\theta(X_i)
	\end{align}
	for some function $f$ and $n$ i.i.d. random draws from some distribution $X_i \sim P$.
    They include maximum likelihood estimators, quantile estimators, regression estimators under $L_2$ or some other loss, as well as many neural networks.
    We extend this to obtain an objective $J_n(\theta; t)$ that depends on both the parameter $\theta$ as well as the threshold $t$. Empirical process theory allows us to show, under the appropriate rescaling, this objective asymptotically converges to a Gaussian process as more data is encountered. 
    Crucially, this Gaussian process has continuous path\change{s}. In other words, when treated as a function of $t$, the Gaussian process is a random, continuous function. Thus, convergence of the objective to a continuous function can be used to obtain convergence of the estimators. 
    
    We are particularly interested in the case where the sample size grows sub-linearly with the data. Here we define an appropriate scaling of the thresholds and show that regardless of how many parameters are in a priority distribution, if all priorities are non-negative and the priority density is \change{sufficiently smooth and} non-zero around 0, the resulting threshold sampler is asymptotically equivalent to a simple weighted sampling procedure where the priority distribution are just parameterized by a univariate weight.
    This means any priority distribution can replicated using $R_i \sim \mathit{Uniform}(0,1/w_i)$ as the priority distribution and appropriate choices of weights $w_i$. This results in an asymptotically equivalent sampling distribution.
    This convergence can only be proved when the class of functions $\{ f_\theta \}_\theta$ used in the objective \change{as well as the collection of thresholding functions}
    \change{are} not too complex. Heuristically, this means they cannot be chosen to overfit the data too much. 
    
    Together these \change{provide} our main results. 
    Loosely speaking, the first states that if (1) an estimator $\hat{\theta}_t$ on deterministic thresholds is consistent and (2) an adaptive threshold $T$ converges to a deterministic threshold in an appropriate asymptotic regime and is not overly complex, then the estimator $\hat{\theta}_T$ on the adaptive threshold is also consistent.
    The second implies that if an adaptive threshold sample grows sub-linearly with the data and priorities are non-negative, the precise choice of priority distribution does not matter. Any priority distribution is asymptotically equivalent to using $R_i \sim \mathit{Uniform}(0, 1/w_i)$ for an appropriate set of weights $w_i$.

	\subsection{Technical Overview}
	We now present an overview \change{of} the technical details. 
	We consider \change{an} extension of an objective function \change{which} includes a threshold as a parameter.
	For any priority distribution and fixed threshold $t$, it is easy to see that 
	the HT-estimator of \change{an M-estimator's} objective \change{$\E_n f_\theta(X_i)$}  can be expressed as an empirical expectation after reweighting by
	\begin{align}
	\hat{J}_n(\theta; t) &= \E_n f_\theta(X_i) w(R, t(X_i)) \quad \mbox{where}\\
	w(R, t(X_i)) &= \frac{1(R_i < t(X_i))}{F_i(t(X_i))} \nonumber.
	\end{align}
	Since our asymptotic results \change{require} an infinite sequence of points, we assume the points $X_i$ are i.i.d. draws from some distribution $P_x$. The corresponding priority $R_i \sim F(\cdot | X_i)$ is drawn from some conditional distribution that depends only on $X_i$. 
	The threshold $t(X_i)$ is also a function of a data point.

	We show that the suitably rescaled objective converges to a Gaussian process, but rather than being indexed by just the parameter of interest $\theta$, it is indexed by both $\theta$ and the threshold $t$. This convergence holds when both the function class $\{f_\theta\}_\theta$ and the class of thresholds $\tcal$ are not too complex.
	
	This allows us to prove consistency of M-estimators under adaptive thresholding schemes. If a threshold $T_n$ converges in probability to a deterministic threshold $t$, 
	then the objective under the random threshold $\hat{J}(\theta; T)$ and fixed threshold
    $\hat{J}(\theta; t)$ converge to the same limit under the continuity of Gaussian processes. Hence, if an estimator is consistent under the deterministic threshold, it is also consistent under the random threshold.
	
	Note that the threshold in this case need not be substitutable, nor does one need to be able to recalibrate it. Heuristic thresholding rules may be used as long as they converge to an appropriate limit. 
	
	We consider two asymptotic regimes, one where the adaptive thresholds converge to fixed thresholds and sample sizes grow linearly with the data, and one where the sample sizes grow sublinearly.
	The first case is straightforward and relies only on the closure properties of bounded uniform entropy integral function classes to prove a Donsker result that $n^{-1/2}(J_n(\theta, t) - J(\theta)) \convp GP_{\theta, t}$ \change{for an appropriate Gaussian process $GP_{\theta, t}$.}
	For the latter \change{case where the sample size grows sublinearly}, we not only need to rescale the objective but also the thresholds to obtain convergence. 
	In this regime, we show that the shape of the priority distribution is asymptotically irrelevant. The asymptotic distribution is only affected by the scaling of priorities. 
	Given convergence of the objective, we can prove our main result 
	\begin{theorem} 
	Let $\hat{\theta}$ be an M-estimator for $\theta_0$ under some distribution $P_{\theta_0}$,
	and suppose its objective $J_n(\theta) = \E_n f_\theta(X_i)$ satisfies the conditions of the M-estimator consistency theorem 2.12 in \cite{kosorok2008empiricalprocesses}.
	Suppose there is a sequence of constants $c_n \to c_0 \geq 0$
    such that $c_n n \to \infty$ and thresholds $T^{(n)} \in \tcal$ with $c_n T^{(n)} \convp T$. If $\{ f_\theta\}_\theta$ and $\tcal$ satisfy the conditions of theorem \ref{thm:sublinear donsker} then the HT-estimate of the objective $\hat{J}_n(\theta, T^{(n)})$ yields a consistent estimator $\hat{\theta}^{(n)}_{T^{(n)}}$ of $\theta_0$.
	\end{theorem}
	\begin{proof}
	    Under these assumptions, theorems \ref{thm:sublinear donsker} and \ref{thm:donsker} and the continuous mapping theorem show that the HT-estimate of the objective also satisfies the the conditions of the M-estimator consistency theorem. Hence, $\hat{\theta}^{(n)}_{T^{(n)}}$ is consistent.
	\end{proof}

    The basic setup for proving convergence to an empirical process starts with a class of functions $\fcal$ and an empirical measure $\mathbb{P}_n$ for $n$ random draws from some measure $P$. This empirical measure takes a function $f \in \fcal$ and maps it to the empirical mean
    \begin{align}
        \mathbb{P}_n f &= \E_{\mathbb{P}_n} f(R_0) = n^{-1} = \sum_{i=1}^n f(R_i)
    \end{align}
    where $R_i \sim P$.
    Since this is a mean, under mild regularity conditions, for any single\change{, real-valued} function $f$, the empirical mean $\mathbb{P}_n f$ converges to a Gaussian random variable by the central limit theorem. 
    Empirical processes theory allows one to show that, if the class of functions $\fcal$ is not too complex, \change{then $\{\mathbb{P}_n f\}_{f \in \fcal}$ converges to a Gaussian \emph{process}.}

	\section{Empirical processes}
	We briefly review some theory for empirical processes to unfamiliar readers. We refer interested readers to \cite{kosorok2008empiricalprocesses} for more details.
	Our goal is to show for a class of functions $\Phi = \{f_\theta\}_\theta$ and threshold functions $\tcal$,
	$\sqrt{n} \left(\hat{J}_n(\theta, t) - J(\theta) \right)$ converges weakly to a Gaussian process $\Psi_{\theta, t}$ 
	with $\cov (\Psi_{\theta,t}, \Psi_{\theta',t'}) = \cov(f_\theta(X) w_t(R,X), f_{\theta'}(X) w_{t'}(R,X))$ for all $f_\theta, f_{\theta'}$ and $t,t' \in \tcal$
	as $n \to \infty$.
	We can write this more succinctly as $\sqrt{n} \left(\hat{J}_n(\theta, t) - J(\theta) \right) \convd \Psi_{\phi, t}$ in $\ell^{\infty}(\Phi \times \tcal)$.	

	Proving this convergence can be done in two steps. First, the finite dimensional distributions and the covariance can be shown to be Gaussian using the usual central limit theorem.
	Second, the empirical process $\Psi^{(n)}_{\phi, t} \defn \sqrt{n} \left(\hat{J}_n(\theta, t) - J(\theta) \right)$
	is shown to be asymptotically tight. Asymptotic tightness ensures that the sample paths of the process are appropriately smooth. Asymptotic tightness is the main challenge for establishing Donsker results.

	In empirical process theory, a sufficient condition for proving asymptotic tightness is that the function class of interest
	is not too complex. There are multiple measures of complexity such as the VC-dimension, bracketing entropy , and uniform entropy. The corresponding conditions that ensure asymptotic tightness are
	finite VC-dimension, finite bracketing entropy integral, and bounded uniform entropy integral with integrable envelope.
	For each of these measures of complexity, there are known, broad classes of functions that satisfy these conditions.
	Donsker preservation theorems show that certain transformations of these classes preserve the conditions on the complexity.
	New classes of interest can often be verified to be Donsker by showing they are contained in a class that is appropriately transformed from these broad, base classes.

	Of these conditions, we are most interested in function classes with finite VC-dimension, also known as a VC-class,
	and those with bounded uniform entropy integral with integrable envelope, known as a BUEI-class with integrable envelope. 
	The envelope $G$ for a class $\gcal$ is the function such that $G(x) \defn \sup_{g \in \gcal} |g(x)|$.
	VC-dimension is the most restrictive of the notions of complexity. 
	Any VC-class is also a BUEI-class. As the most restrictive measure of complexity, it allows for the greatest range of transformations. In particular, it allows for composition with monotone functions. BUEI-classes are of interest because a product of BUEI-classes remains a BUEI-class.

	\subsection{Donsker result for fixed thresholds}
    We first consider the case where an adaptive threshold converges to a fixed one, and
    the sample size grows linearly with the data. 

	\begin{theorem}
		\label{thm:donsker}
		Let $\Phi$ be a BUEI-class of functions on a data point, and let $\tcal$ be a class of threshold functions that has finite VC-dimension. Further assume that $F(t(x)) > \epsilon$ for some $\epsilon > 0 $ and all $x \in \xcal$.  
		Then, $f_\theta(X) w(R, t(X))$
		has bounded uniform entropy integral. Hence, 
		\begin{align}
			\hat{J}_n(\theta,t) &= \E_n f_\theta(X) w(R, t(X)) \\
			\sqrt{n}(\hat{J}_n(\theta,t) - J(\theta)) &\convd \Psi^0_{\theta,t}
		\end{align}
		where $\Psi^0_{\theta, t}$ is a Gaussian process
		indexed by $f_\theta \in \Phi$ and $t \in \tcal$.
	\end{theorem}
	\begin{proof}
		Since $\tcal$ has finite VC-dimension, the composition rules for VC-classes (lemma 9.9 in \cite{kosorok2008empiricalprocesses}) 
		give that both 
		$1(r - t(x) < 0)$ and $F(t(x))$ generate VC-classes since the indicator and any CDF are monotone.
		Likewise, $1/F(t(x))$ is a VC-class. Furthermore, it has a measurable envelope $H(x) = 1/\epsilon$ \change{since $F(t(x)) > \epsilon$ by assumption}.
		Since VC-classes \change{are} BUEI\change{-classes} if there exists a envelope, and BUEI classes are closed under multiplication (theorem 9.15 in \cite{kosorok2008empiricalprocesses}), the class of functions $\{g(x) w_t(r,x) : g \in \gcal, t \in \tcal\}$ \change{is a} BUEI\change{-class} with envelope $G\cdot H$. 
		This establishes asymptotic tightness of the process $\Theta = \{ {\theta}_{gt} \}_{g,t}$
		\change{where $\theta_{gt}$ is the Horvitz-Thompson estimator for $\E g$ when using a threshold $t$}.
		Since Donsker classes are preserved under multiplication by a bounded, measurable function (corollary 9.31 in \cite{kosorok2008empiricalprocesses}), 
		$\gcal \cdot w_t$ is $P$-Donsker for any fixed $t$.
		Since \change{$P$-Donsker classes} are also closed under finite sums, 
		the finite dimensional distributions of $\Theta$
		converge to multivariate Gaussian distributions.
		Asymptotic tightness and the central limit theorem on finite dimensional distributions imply
		that $\Theta$ is $P$-Donsker.
	\end{proof}

	\subsection{Donsker result on sub-linear samples}
	In the above result, the resulting sample sizes must grow linearly with the data. We are interested in the case where the sample sizes still grows to infinity, but the inclusion probabilities go to 0. 
	We show that under appropriate conditions, this is sufficiently well-approximated by a two-step procedure which downsamples the uniformly to obtain a sublinear number of data points and then applies threshold sampling where the inclusion probabilities are bounded away from 0 as above.

	\begin{theorem}
	\label{thm:sublinear donsker}
	Consider priorities $R_i$ taking values in the non-negative reals.
	Further suppose their conditional CDFs $F(\cdot | x)$ have a linear expansion near 0
	\begin{align}
		\Delta(r) &\defn \sup_{x \in Supp(P_x)} |F(r|x) - w_x r| = \change{o(r)} \quad \mbox{if $r \geq 0 $}
	\end{align}
	for some weights $w_x$ for all $x \in Supp(P_x)$ such that $M \defn \sup_x w_x < \infty$ and $\inf_x w_x > 0$.
	
	Let $\Phi$ be a class of functions $\Phi = \{f_\theta\}$ and $\tcal$ a class of thresholds $\tcal$ satisfying the conditions of theorem \ref{thm:donsker}. Further assume that the classes are uniformly bounded with $\lVert t \rVert _{\infty}< T$ for all $t\in \tcal$.
	Furthermore, consider alternative priorities $\dot{R}_i |X_i = x \sim \mathit{Uniform}(0, 1/w_x)$
	and let $\change{\dot{J}}_n(\theta, t)$ be the estimated objective using these priorities.
	
	Then, for a sequence $c_n \to 0$ such that $c_n n \to \infty$,
	the processes
	\begin{align}
	\label{eqn:G}
		G_{\theta, t}^{(n)} &\defn (c_n M T n)^{1/2} \left(\hat{J}_n(\theta, c_n t) - J(\theta)\right) \convd \Psi_{\theta, t} \\
	    \dot{G}_{\theta, t}^{(n)} &\defn n^{1/2} \left(\dot{J}_n(\theta, t / MT) - J(\theta)\right)
	    \convd \dot{\Psi}_{\theta, t}
	    \label{eqn:Gdot}
	\end{align}
	as $n \to \infty$
	where $\Psi_{\theta,t}, \dot{\Psi}_{\theta, t}$ are Gaussian processes
	and $\Psi_{\theta,t} \eqdist \dot{\Psi}_{\theta, t}$.

	\end{theorem}

	We first outline the proof. Rather than dealing with a fixed function class as before, these conditions require proving a Donsker result when the class of thresholds is changing with $n$. Bounded uniform entropy conditions also exist for this case which ensure convergence to a Gaussian process. However, when directly applied, the decreasing thresholds lead to envelopes whose integrals go to $\infty$.
	To handle that, we show that the random process can be generated in two stages: a uniform sampling stage that draws a sample with sublinear size and on the order of $n c_n$ and a second stage where the thresholds do not go to 0. 
	The boundedness conditions ensure Donsker preservation results can be applied on this second stage and obtain convergence to a Gaussian process. 
	We then compare the process using priorities $R_i$ with the approximating process using $\dot{R}_i$ and show their first two moments converge to the same limit. Hence, their Gaussian process limits must also be the same.

	\begin{proof}		
        First, note that for any threshold $t$, inclusion of an item $X_i$ is equivalent to 
        $U_i < F(c_n t(X_i) | X_i)$ for some $U_i \sim \mathit{Uniform}(0,1)$.
		Since $F(c_n t(x) | x) = w_x c_n t(x) + o(w_x c_n t(x)) < c_n M T + c_n \epsilon$ eventually for any $\epsilon > 0$,
		this inclusion event can be rewritten as \change{two steps. First, } $U_i < c_n \gamma_{\epsilon}$ \change{where $\gamma_\epsilon \defn M T + \epsilon$} and, if it passes this first stage, 
		$\frac{U_i}{c_n \gamma_{\epsilon}} < \frac{F(c_n t(x) | x)}{c_n  \gamma_{\epsilon}}$. Note that this second stage uses the conditional distribution
		$\frac{U_i}{c_n \gamma_{\epsilon}} | U_i < c_n \gamma_{\epsilon} \sim \mathit{Uniform}(0,1)$. 
		The first stage of the inclusion event is a uniform Poisson sample. It thins the data by drawing i.i.d. $Bernoulli(c_n \gamma_{\epsilon})$ draws. The resulting sample of size $C_n$ is still from the same base distribution $P$ as the original data.
		Thus, we have that
		\begin{align*}
			\hat{J}_n(\theta, c_n t) &= \E_n f_\theta(X) \frac{1(U < F(c_n t(X)|X))}{F(c_n t(X) | X)} \\
			&= \frac{C_n}{n} \frac{1}{c_n \gamma_{\epsilon}} \E_{C_n} f_\theta(X) \frac{1\left(U < \frac{F(c_n t(X)|X)}{c_n \gamma_\epsilon}\right)}{\frac{F(c_n t(X)|X))}{c_n \gamma_\epsilon}}.
		\end{align*} 
		Since $C_n \sim Binomial(n, c_n \gamma_\epsilon)$ and $n c_n \to \infty$, it follows that  $\frac{C_n}{n} \frac{1}{c_n \gamma_{\epsilon}} \convp 1$
		and can be ignored by Slutzky's lemma. 
		
		Thus, we can instead examine the convergence of the process $\E_{C_n} \phi(X) \frac{1\left(U < F(c_n t(X)|X) / c_n \gamma_\epsilon\right)}{F(c_n t(X)|X)) / c_n \gamma_\epsilon}$. Let $\vcal$ be the VC-index of $\tcal$. The threshold 
		$\{ F(c_n t(X)|X)) / c_n \gamma_\epsilon\,:\, t\in \tcal \}$ is a VC-class with index  $\leq \vcal$ since scalar transformations do not change the VC-index and monotone transformations composed with a VC-class do not increase the VC-index. 
		Furthermore, since $F(c_n t(X)|X=x)) / c_n \gamma_\epsilon \to w_x t(x) / M T \leq 1$ 
		and $\Phi$ is a BUEI-class that is uniformly bounded by some constant $H$, the empirical expectations are taken over  function classes that are uniformly bounded by $H$. 
		Together these imply  
		\vspace{-0.1cm}
		\begin{align}
		\label{eqn:process}
		\sqrt{C_n} \left(\E_{C_n} f_\theta(X) \frac{1\left(U < \frac{F(c_n t(X)|X)}{c_n \gamma_\epsilon}\right)}{\frac{F(c_n t(X)|X))}{c_n \gamma_\epsilon}} - J(\theta) \right) \end{align}
        converges to a Gaussian process limit with mean 0 as long as its finite dimensional marginals 
		converge to multivariate Gaussians. The law of large numbers gives that $n c_n \gamma_\epsilon / C_n \to 1$. Since $\epsilon$ can be arbitrarily small, equation \ref{eqn:G} is proved.
		
		Now consider the process which replaces $F(r|x)$ with the approximation $\dot{F}(r | x) = w_x r$. 
		We wish to show that the mean and covariances of the process in equation \ref{eqn:process} remain the same after the substitution.
		Since the HT-estimator is unbiased regardless of the choice of $F$, the mean is 0. Let $Z_n = 1\left(U < \frac{F(c_n t(X)|X)}{c_n \gamma_\epsilon}\right)$
		and $\dot{Z}_n$ be the same with $F$ replaced by $\dot{F}$.
		
	    Note, $F(c_n r|x) = c_n w_x r + \change{o(c_n w_x r)}$.
		The difference in inclusion variables weighted by \change{the inverse} pseudo-inclusion probability is %
		\vspace{-0.1cm}
		\begin{align*}
			\Delta_n &= \frac{Z_n}{F(c_n t(X)|X)) / c_n \gamma_\epsilon }  - \frac{\dot{Z}_n}{\dot{F}(c_n t(X)|X)) / c_n \gamma_\epsilon} \\
			&= \frac{\gamma_\epsilon}{w_X t(X)}  \left(Z_n (1+o(t(X)))  - \dot{Z}_n\right) 
		\end{align*}
		This gives $|\Delta_n| \leq \change{\frac{\gamma_\epsilon}{w_X t(X)} \change{o(1)}}$ 
		if $Z_n = \dot{Z}_n$ and $|\Delta_n| \leq \change{\frac{\gamma_\epsilon}{ w_X t(X)}(1 + o(1))}$ otherwise.
		Since $P(Z_n \neq \dot{Z}_n) = \change{o(1)}$ \change{and both $w_X$ and $t(X)$ are bounded away from 0}, %
		the variance $Var(\Delta_n) = \change{o(1)} \to 0$ as $n \to \infty$.
		Since $\Phi$ has integrable envelope, $\|f_\theta\|^2$ is bounded.
		The Cauchy-Schwartz inequality gives that covariances of the processes using $F$ and $\dot{F}$ are equal.
		Thus, $\hat{J}$ in equation \ref{eqn:G} can be replaced by \change{$\dot{J}$}
		while still converging to the same limit.
		Finally, note that when priorities are from $\dot{F}$, the two-stage sampling trick provides a means to rescale the threshold. This gives
		$\change{\dot{J}}_n(\theta, \alpha \delta t) \eqdist \change{\dot{J}}_B(\theta, \delta t)$
		where $B \sim Binomial(n, \alpha)$. Using $\dot{F}$ in place of $F$ and rescaling the threshold in equation \ref{eqn:G} by $c_n M T$ yields equation \ref{eqn:Gdot}.

		In the thinning stage, the role of $\epsilon$ is simply to ensure that $F(c_n t(X) | X)$ can be upper bounded by $c_n (M T + \epsilon)$. Any value of $\epsilon$ that yields an upper bound yields exactly the same process in the end. Since $\dot{F}(c_n t(X) | X) \leq M T$ already, $\epsilon$  can be set to 0.

	\end{proof}

Although the condition on the CDFs requiring a linear expansion near 0 may appear restrictive, we note that it is satisfied under reasonable settings. If the priorities are drawn from conditional densities $f(\cdot | X)$ that are differentiable in a neighborhood $[0, \delta)$ with $\delta > 0$ and if $f(0 | \cdot)$ is both upper bounded and bounded away from 0, then a Taylor expansion ensures the condition is satisfied. 
\change{For example, if $f(\cdot | x)$ is the $\mathit{Uniform}(0, 1/x)$ density associated with priority sampling or an $Exponential(x)$ density, then the condition is satisfied if $x$ can be bounded.} Furthermore, 
\change{
even if the original CDF does not have a linear expansion, a monotone transformation of the priorities may be able to convert the priorities into ones where} the CDF has a linear expansion. \change{The following lemma shows that if the ratio of CDFs for priorities has a limit at 0
then they are asymptotically equivalent to using $\mathit{Uniform}$ priorities, where the CDFs do have a linear expansion.}

\begin{lemma}
	\label{lem:priorities to weights}
	Consider priorities $R_i$ taking values in the non-negative reals. 
	Further suppose their conditional distributions $F(\cdot | x)$ are continuous in a neighborhood of $0$ with $F(0 | x) = 0$
	and there exists constants $w_x$ with $\inf_x w_x > 0$ such that for $\delta \to 0^+$ %
	\begin{align}
		\sup_{x,y} \left| \frac{F(\delta | x)}{F(\delta|y)} - \frac{w_x}{w_y} \right| = o(1).
	\end{align}
	There exist independently drawn priorities with conditional distribution $\dot{R}_i | X_i = x \sim \mathit{Uniform}(0, 1/w_x)$ 
	and a monotone transformation $\rho$ 
	such that $p( 1(\dot{R}_i < t) \neq 1(\rho(R_i) < t)) = o(t)$.
\end{lemma}
\begin{proof}
	The conditions imply $\sup_x |F(\delta | x) - \alpha_x \eta(\delta)| = o(\eta(\delta))$ by taking $\eta(\delta) = \alpha_y^{-1} F(\delta | y)$ for some fixed $y$.	
	Since $\eta$ is increasing and continuous in a neighborhood of $0$, for sufficiently small $b$ we can  simply rescale the priorities to obtain $\tilde{R}_i = b \eta^{-1}(R_i)$ if $R_i \in [0,b)$ and $\tilde{R}_i = R_i$ otherwise. This defines the function $\rho$.
	Denote the CDF of a $\mathit{Uniform}(0,1/w_x)$ distribution as $\dot{F}(r | x) = w_x r$.
	The priorities $\tilde{R}_i = \tilde{F}^{-1}(U_i | X_i)$ and $\dot{R}_i = \dot{F}^{-1}(U_i | X_i)$ can be obtained from the inverse probability transform for their corresponding CDFs.
	Thus, the indicators are not equal only if 
	$\dot{F}(t | X_i) \leq U_i < \tilde{F}(t | X_i)$ or $\tilde{F}(t | X_i) \leq U_i < \dot{F}(t | X_i)$.
	Each of these happen with probability less than 
	$\sup_x |\tilde{F}(t | x) - \dot{F}(t | x)| = o(t)$.
\end{proof}

\section{Applications of asymptotic theory}
While we have already shown that the asymptotic theory justifies the re-use of consistent estimators for Poission sampling, we can also use it to justify the use of heuristically constructed thresholds.

Consider again the problem of building a sample that can provide an absolute variance guarantee. Recall that in section \ref{sec:variance sized samples}, it is not sufficient to choose a threshold where the estimated variance is equal to the target variance. Additional points must be sampled to ensure that the chosen threshold is the largest of such thresholds. This added layer of complexity can be removed by applying the asymptotic theory to show that the heuristically constructed threshold without oversampling still leads to consistent estimators. 

Our Donsker result show that the HT variance estimate in \ref{eqn:HT var estimate} after centering and rescaling
$\widehat{Var}(\hat{\theta}_t) \approx Var(\hat{\theta}_t)) + \Psi_t/\sqrt{n}$ where $\Psi_t$ is a zero-mean Gaussian process.
Since $Var(\hat{\theta}_t))$ is increasing with $t$ and 
the error term $\Psi_t / \sqrt{n}$ is decreasing with $n$,
if the variance estimator's variance is not too large,
then a Gaussian maximal inequality can be used to show the heuristically chosen threshold without the additional oversampling step is close to the threshold with the desired variance.

	\section{Future work and Conclusion}
    We have provided a general framework for building sampling schemes that adapt to the data on the fly and satisfy system constraints while behaving similarly to fixed thresholds. This simplifies creating new sampling schemes while making the resulting sampled data sets easy to analyze. This framework unifies a long line of research on bottom-k sampling. 
    
    We demonstrate its usefulness by providing sampling schemes that can solve a variety of existing and new problems. These include engineering problems of fitting within system and budgetary constraints, usability problems that allow users to tune the desired accuracy for AQP results at query time, and novel problems such as top-k queries. This flexibility in the framework can make it more useful in practice, by making it easier to design systems and satisfy user requirements. At the same time, it ensures that the same estimators used for fixed thresholds can also be used for adaptive thresholds, making it easy to code just one set of estimators while the underlying sampling schemes can be easily changed.
    Furthermore, these are just a few examples that apply the framework. 
    Future work expands on other potential applications that can be solved with this sampling framework.

	\bibliographystyle{abbrv}
	\balance
	\bibliography{ling2}
	
\end{document}